\documentclass[journal,twoside,web]{ieeecolor}

\usepackage{silence}
\usepackage[font=scriptsize]{caption}
\usepackage{array}
\usepackage{dirtytalk}
\usepackage{algorithm} 
\usepackage{algpseudocode} 
\usepackage{soul}
\usepackage{booktabs, makecell, multirow, tabularx}

\usepackage{generic}
\usepackage{amsmath,amssymb,amsfonts}

\usepackage{graphicx}
\usepackage{textcomp}
\usepackage{booktabs}
\usepackage{multirow}
\usepackage{hyperref}

\usepackage{enumerate}
\usepackage{pifont}
\newcommand{\xmark}{\ding{55}}%
\newcolumntype{C}{>{\centering\arraybackslash}X}
\usepackage{fancyhdr}
\fancypagestyle{FirstPage}{

\lhead{\textcolor{red}{$\copyright$ 2024 IEEE.  Personal use of this material is permitted.  Permission from IEEE must be obtained for all other uses, in any current or future media, including reprinting/republishing this material for advertising or promotional purposes, creating new collective works, for resale or redistribution to servers or lists, or reuse of any copyrighted component of this work in other works}}
\lfoot{Corresponding E-mail: usmanm@stanford.edu}

}

\def\BibTeX{{\rm B\kern-.05em{\sc i\kern-.025em b}\kern-.08em

    T\kern-.1667em\lower.7ex\hbox{E}\kern-.125emX}}

\begin{document}


\title{Multi-Task Adversarial Variational Autoencoder for Estimating Biological Brain Age with Multimodal Neuroimaging}

\author{Muhammad~Usman, Azka~Rehman, Abdullah~Shahid, Abd~Ur~Rehman, Sung-Min~Gho, Aleum~Lee, Tariq~M.~Khan, and Imran~Razzak, \IEEEmembership{Senior Member, IEEE} } 

\maketitle

\begin{abstract}
Despite advances in deep learning for estimating brain age from structural MRI (sMRI), incorporating functional MRI (fMRI) data presents significant challenges due to its complex data structure and the noisy nature of functional connectivity measurements. To address these challenges, we present the Multitask Adversarial Variational Autoencoder (M-AVAE), a bespoke deep learning framework designed to enhance brain age predictions through multimodal MRI data integration. The M-AVAE uniquely separates latent variables into generic and unique codes, effectively isolating shared and modality-specific features. Additionally, integrating multitask learning with sex classification as a supplementary task enables the model to account for sex-specific aging nuances. Evaluated on the OpenBHB dataset—a comprehensive multisite brain MRI aggregation—the M-AVAE demonstrates exceptional performance, achieving a mean absolute error of 2.77 years, surpassing conventional methodologies. This success positions M-AVAE as a powerful tool for metaverse-based healthcare applications in brain age estimation. The source code is made publicly available at: \href{https://github.com/engrussman/MAVAE}{https://github.com/engrussman/MAVAE}.
\end{abstract}


\section*{Introduction}
\label{intro}
\thispagestyle{FirstPage}

The advent of multimodal neuroimaging, which combines functional magnetic resonance imaging (fMRI) for the assessment of functional connectivity and structural magnetic resonance imaging (sMRI) for cortical morphology, offers a nuanced approach to the detection of cognitive impairment and the prediction of brain age \cite{bib_9}. However, the exploration of anatomical and functional differences in the brain between sexes using multimodal imaging for the estimation of brain age remains underexplored.


Sex differences play a vital role in the brain's ageing process, with notable anatomical and functional variations between male and female brains \cite{franke2013gender}. Incorporating sex information into age estimation models improves accuracy and has shown promise in deep learning applications \cite{sanford2022sex}. Our research addresses this gap by integrating sex considerations in a multimodal imaging framework within the metaverse context, aiming to improve the accuracy and applicability of brain age predictions in personalised healthcare.

Specifically, we propose a novel metaverse-based AI application for brain age estimation: the Multi-Task Adversarial Variational Autoencoder (M-AVAE). This innovative model merges adversarial learning and variational auto-encoding capabilities within a multitask learning framework, aiming for simultaneous estimation of brain age and prediction of sex from multimodal MRI data, including both sMRI and fMRI. The design of M-AVAE meticulously segregates the latent features of each imaging modality into distinct components, effectively disentangling the shared and unique attributes across modalities. This method not only improves the accuracy in capturing commonalities, but also minimises interference during data fusion, presenting a novel approach to multimodal neuroimaging analysis suitable for integration into metaverse platforms.

Our major contributions can be summarised as follows.

\begin{itemize}
    \item We introduce a novel multimodal framework for the estimation of brain age within the metaverse ecosystem for healthcare. Integrating our AI model into a metaverse environment, may enable continuous, real-time updates and interactions, enhancing the precision and reliability of brain age estimations and may allow personalised, predictive healthcare.
    \item Our approach is unique in creating a disentangled representation of brain imaging data by applying both adversarial and variational principles within a single architecture. This disentanglement allows for the clear differentiation of shared versus modality-specific information, paving the way for more nuanced interpretations of neuroimaging data within a metaverse platform. 
    \item Through rigorous evaluation of publicly available datasets, our extensive experiments validate the efficacy and robustness of the proposed framework, thereby establishing a new benchmark for brain age estimation models suitable for metaverse integration.
\end{itemize}

\begin{table}[ht]
\centering
\caption{Summary of comparison of our work with the existing studies in term of availability of different components, i.e., multimodal, multitask, adversarial, and distangled learning.}
\label{litrature_tab}
\begin{tabularx}{\columnwidth}{@{}lCCCCC@{}}
\toprule
\textbf{\scriptsize Author} & \textbf{\scriptsize Multimodal} & \textbf{\scriptsize Multitask}  & \textbf{\scriptsize Adversarial} & \textbf{\scriptsize Disentangled}\\
\textbf{\scriptsize (Year)} & \textbf{\scriptsize learning} & \textbf{\scriptsize learning} & \textbf{\scriptsize learning} & \textbf{\scriptsize learning} \\
\midrule
{\scriptsize He \textit{et al.} 2022 \cite{he2022global}} & \xmark & \xmark & \xmark & \xmark\\ 
{\scriptsize He \textit{et al.} 2022 \cite{he2022deep}} & \xmark & \xmark & \xmark & \xmark\\ 
{\scriptsize Cheng \textit{et al.} 2021 \cite{cheng2021brain}} & \xmark & \xmark & \xmark & \xmark\\ 

{\scriptsize Armanious \textit{et al.} 2021 \cite{armanious2021age}} & \xmark & \checkmark & \xmark & \xmark\\ 
{\scriptsize Zhang \textit{et al.} 2022 \cite{zhang2022robust}} & \xmark & \checkmark & \xmark & \xmark\\ 
{\scriptsize Liu \textit{et al.} 2023 \cite{liu2023risk}} & \xmark & \checkmark & \xmark & \xmark\\ 
{\scriptsize Dular \textit{et al.} 2024 \cite{dular2024base}} & \xmark & \checkmark & \xmark & \xmark\\ 
{\scriptsize Wang \textit{et al.} 2023 \cite{wang2023skewed}} & \xmark & \checkmark & \xmark & \xmark\\ 

{\scriptsize Mouches \textit{et al.} 2022 \cite{mouches2022multimodal}} & \checkmark & \checkmark & \xmark & \xmark\\
{\scriptsize Cai \textit{et al.} 2023 \cite{cai2023graph}} & \checkmark & \xmark & \xmark & \checkmark \\ 
{\scriptsize Hu \textit{et al.} 2020 \cite{hu2020disentangled}} & \checkmark & \xmark & \checkmark &\checkmark\\

\textbf{\scriptsize Our Study} & \checkmark & \checkmark & \checkmark   & \checkmark \\ 
\bottomrule
\end{tabularx}
\end{table}

\section*{Related Work}
\label{rw}
Most studies on brain age estimation have utilised T1-weighted sMRI scans \cite{tanveer2023deep}, while recent research highlights the potential of functional magnetic resonance imaging (fMRI) to predict cognitive variables \cite{bedel2023bolt, bedel2023dreamr}. The ability of fMRI to capture intricate patterns of brain activity makes it valuable for the prediction of brain age. Several studies have explored multimodal MRI to improve prediction accuracy \cite{jirsaraie2023systematic}. Multimodal data fusion techniques are typically classified as model-agnostic or model-based \cite{bib_20}. Model-agnostic fusion includes early fusion (EF), where features of multiple modalities are combined as input to a single model, and late fusion (LF), which integrates decision values using mechanisms such as averaging or voting \cite{liu2018learn}. These methods often under-use cross-modal correlations, as noted in prior studies on MRI-based image enhancement and classification \cite{ullah2020hybrid}. In contrast, model-based fusion techniques, such as multiple kernel learning, graphical models, and neural networks, aim to build generalised models. Studies utilizing attention-based and transformer networks for improved segmentation, particularly in challenging modalities, have shown promise in advancing model-based approaches \cite{farooq2023residual, kanwal2023mask, usman2024intelligent,usman2023mesaha, usman2024meds, rehman2023selective, iqbal2023ldmres, usman2023deha, usman2022dual, ullah2023ssmd, latif2018automating, lee2021evaluation, latif2018mobile, ullah2023mtss}. Autoencoders (AE) have been widely explored for multimodal fusion, with early and late fusion \cite{bib_20, farooq2023dc, latif2020leveraging, ullah2023densely, usman2017using, ullah2022cascade, usman2020retrospective}. However, traditional AEs often struggle to differentiate between shared and complementary information, and noisy modalities can negatively impact latent representation learning across modalities \cite{usman2020volumetric, latif2018cross, latif2018phonocardiographic}. 

A concise summary of the existing literature is provided in Table \ref{litrature_tab}. Our analysis highlights key methodologies, emphasising the role of modality, multitasking, adversarial learning, and disentangled autoencoders. For example, He et al. \cite{he2022global} employed a global-local transformer architecture for age estimation, integrating global and local information via an attention mechanism. In another study \cite{he2022deep}, the same group introduced deep relation learning for regression, designed to capture relationships between pairs of input images. They utilised an efficient convolutional neural network (CNN) to extract features from these image pairs and a transformer for relation learning. However, this approach did not incorporate gender information and relied solely on single-modality MRI scans. Similarly, Cheng et al. \cite{cheng2021brain} proposed a two-stage cascade network that used MRI images and gender labels, employing ranking losses for age prediction. However, these studies \cite{he2022global,he2022deep,cheng2021brain} focused mainly on single-modality MRI scans and lacked gender integration, a critical factor in accurate age prediction.

Moreover, Zhang et al. \cite{zhang2022robust} introduced an adaptive ensemble learning framework for robust age estimation on unimodal sMRI data, categorising subjects into different age groups and sex, and selected the most suitable ensemble model. Similarly, Liu et al. \cite{liu2023risk} and Armanious \cite{armanious2021age} proposed methods to estimate brain age using Support Vector Regression (SVR) and a deep CNN, respectively. Both studies incorporated gender information into their models and emphasised the discrepancy between chronological and biological ages. Recently, Dular et al. \cite{dular2024base} developed a deep learning model trained on multisite T1-weighted MRI data to predict age and age class. Wang et al. \cite{wang2023skewed} addressed predictive bias in brain age estimation by employing a skewed loss function. However, similar to previous studies \cite{zhang2022robust, liu2023risk, armanious2021age, dular2024base}, this approach did not utilise multimodal data.

A few studies \cite{mouches2022multimodal, cai2023graph, hu2020disentangled}, as summarised in Table \ref{litrature_tab}, have explored the use of multimodal data for the prediction of brain age. Mouches et al. \cite{mouches2022multimodal} proposed a CNN-based multitasking network combined with a linear regression model to predict the age of the biological brain from multimodal data, also computing saliency maps to localise relevant brain regions. Cai et al. \cite{cai2023graph} introduced a two-stream convolutional autoencoder to separate distinct information from each modality, incorporating disentangled autoencoders, although gender prediction was not included. Similarly, Hu et al. \cite{hu2020disentangled} developed a disentangled multimodal adversarial network for age prediction, focussing solely on male subjects. Although multi-encoder architectures with various types of loss functions have shown significant performance improvements \cite{usman2024meds, rehman2023selective}, they have not been sufficiently explored for brain age estimation applications.

In contrast to these studies, our work is the first to propose a multimodal architecture that effectively combines adversarial learning with multitask learning to address the gaps in existing research. Using adversarial learning-based disentangled autoencoders, our model provides a structured framework for the estimation of brain age using multimodal data, while also integrating gender classification within a multitask learning framework.

\begin{figure*}[ht]
\centering
\includegraphics[width=0.7\textwidth]{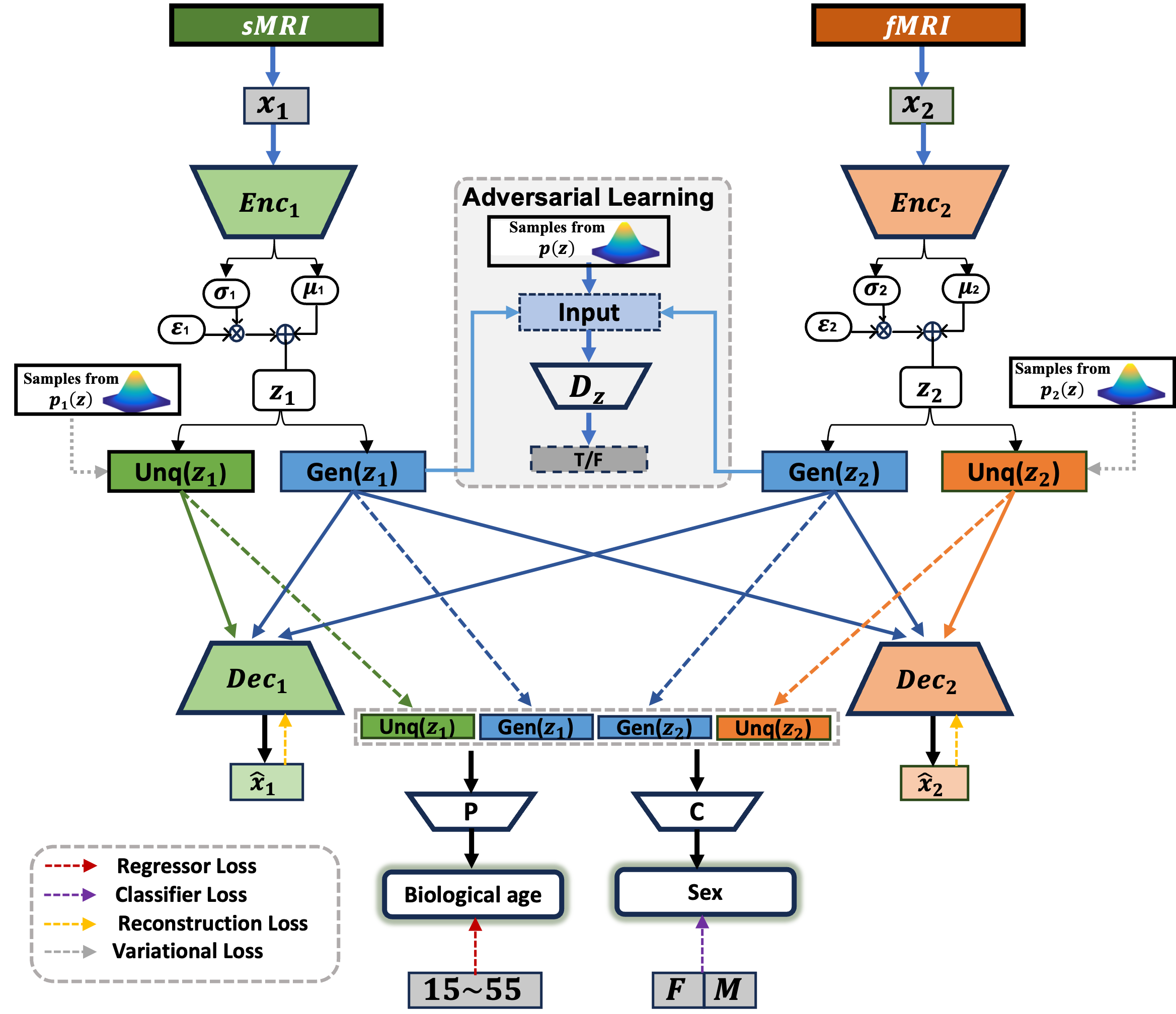}
\caption{Architecture of our proposed Multitask Adversarial Variational Autoencoder (M-AVAE) for predicting brain age and sex estimation from multimodal MRI data which includes sMRI and fMRI scans. }
\label{proposed_architecture}
\end{figure*}

\section*{Proposed Method}
\label{pro}

This section delineates the Multitask Adversarial Variational Autoencoder Network (M-AVAE), which leverages multimodal inputs—specifically sMRI and fMRI scans—to predict biological age accurately. Our approach begins with feature extraction from both modalities, followed by integration into the M-AVAE. The model comprises dual autoencoders and a combination of loss functions that work to optimise the accuracy of brain age estimation. The architecture is depicted in Fig. \ref{proposed_architecture}, and the following subsections break down the components of the proposed model.

\subsection*{Feature Extraction Process}
\label{preprocessing}

Given the high-dimensional nature of MRI data and limited dataset sizes, direct input of these features into neural networks can lead to overfitting. Thus, feature selection is a critical preprocessing step. Feature selection methods are broadly classified into filter, wrapper and embedded methods \cite{urbanowicz2018relief}. Consistent with prior work \cite{hu2020disentangled}, we use the filter method due to its model independence and efficiency. Specifically, the Random Forest algorithm is used to handle highly correlated and high-dimensional data \cite{kawakubo2012rapid}. After selecting the features, we obtain the $m_1$ and $m_2$ features from the two modalities for further analysis. The dataset can be represented as:

\begin{equation}
X_{1n} \in \mathbb{R}^{m_1}, \quad X_{2n} \in \mathbb{R}^{m_2}, \quad y_n \in \mathbb{R},
\end{equation}

where $X_{1n}$ and $X_{2n}$ are feature vectors from the first and second modalities, respectively, and $y_n$ corresponds to the target outputs (age and sex) for $N$ instances.

\subsection{Encoder Branches and Latent Variables}
For each modality, a multilayer perceptron (MLP) serves as the encoder, denoted $Enc_i$ for $i = 1, 2$. The encoder transforms the input feature vectors into latent representations:

\begin{equation}
    z_i = Enc_i(x_i),
\end{equation}

where $z_i$ is divided into a shared component $\text{Gen}(z_i)$ and a unique component $\text{Unq}(z_i)$ to capture the shared and modality-specific information. This separation ensures that the encoder can reconstruct the latent representation of its components while maximising the similarity of shared codes between modalities and distinguishing unique codes.

\subsection{Cross Encoder Reconstruction}

An MLP acts as a decoder $Dec_i$ for each modality. Given $z_i = [\text{Gen}(z_i), \text{Unq}(z_i)]$, the decoder reconstructs $x_i$:

\begin{equation}
x_i' = Dec_i(\text{Gen}(z_i), \text{Unq}(z_i)).
\end{equation}

Additionally, the shared codes $\text{Gen}(z_i)$ are used to reconstruct the inputs from the other modality, ensuring that:

\begin{equation}
x_i' = Dec_i(\text{Gen}(z_j), \text{Unq}(z_i)) \quad \text{for} \quad i \neq j,
\end{equation}

which enforces the separation of shared and unique information.

\subsection{Age and Sex Prediction Strategy}
Disentangling each modality's latent variable into generic $\operatorname{Gen}(\mathbf{Enc}_i(\mathbf{x}_i))$ and unique $\operatorname{Unq}(\mathbf{Enc}_i(\mathbf{x}_i))$ codes, we form $M(\mathbf{x}_1, \mathbf{x}_2)$ as:
\begin{equation}
M(\mathbf{x}_1, \mathbf{x}_2) = \left(\text{Gen}_{1,2}, \operatorname{Unq}(\mathbf{Enc}_1(\mathbf{x}_1)), \operatorname{Unq}(\mathbf{Enc}_2(\mathbf{x}_2))\right),
\end{equation}
where $\operatorname{Gen}_{1,2} = \sum_{i=1}^2 \omega_i \operatorname{Gen}(\mathbf{Enc}_i(\mathbf{x}_i))$ with $\omega_1 = \omega_2 = 0.5$. Two MLPs, a regressor $\mathbf{P}$ and a classifier $\mathbf{C}$, predict age and gender from $M(\mathbf{x}_1, \mathbf{x}_2)$, respectively. Our approach integrates adversarial autoencoders (AAE) \cite{makhzani2015adversarial} with variational loss \cite{kingma2013auto} for each modality, facilitating disentanglement and information fusion in cross-reconstruction. A shared MLP discriminator $\mathbf{D_z}$ enforces adversarial regularisation on $\mathbf{z}_i$, guiding it toward a predefined distribution. The weights of $\mathbf{Enc}_i$, $\mathbf{Dec}_i$, $\mathbf{D_z}$, $\mathbf{P}$, and $\mathbf{C}$ are learnt together with specific loss functions.

\subsection{Objective Function}
\subsubsection{Adversarial Loss}
Let the prior distribution imposed on the generic part of the latent variable be denoted by $p(\mathbf{z})$, the encoding distribution by $q(\mathbf{z}_i | \mathbf{x}_i)$, and the decoding distribution by $p(\mathbf{x}_i | \mathbf{z}_i)$. The Adversarial Autoencoder (AAE) aims to learn the data distribution $p_d(\mathbf{x}_i)$ by training an autoencoder with a regularised latent space. This involves ensuring that the aggregated posterior distribution $q(\mathbf{z}_i) = \int_{\mathbf{x}_i} q(\mathbf{z}_i | \mathbf{x}_i) p_d(\mathbf{x}_i) d\mathbf{x}_i$ matches the predefined prior $p(\mathbf{z}_i)$. Regularisation is achieved through an adversarial process involving the discriminator $\mathbf{D_z}$, leading to a minimax problem:

The total adversarial loss is the sum of the adversarial losses from both modalities:
\begin{equation}
\mathcal{L}_{adv} = \mathcal{L}_{adv}^1 + \mathcal{L}_{adv}^2.
\end{equation}

The encoder's goal is to make the generic part of the posterior distribution indistinguishable by the discriminator $\mathbf{D_z}$ from the prior distribution $p(\mathbf{z}_i)$, while $\mathbf{D_z}$ aims to differentiate between $q(\mathbf{z}_i)$ and $p(\mathbf{z}_i)$. In this study, a Gaussian prior distribution $\mathcal{N}(\mu_i(\mathbf{x}_i), \sigma_i(\mathbf{x}_i))$ is used for $\mathbf{z}_i$, applying the reparameterisation trick for efficient backpropagation.

\subsubsection{Variational Loss}
To regularise unique and generic codes without significantly increasing computational complexity, we integrate the Variational Autoencoder (VAE) approach, known for its ability to impose prior distributions through KL divergence, as an alternative to using multiple discriminator networks. This approach, referred to as variational loss, enforces different prior distributions $p(\mathbf{z}_1)$ and $p(\mathbf{z}_2)$ for the unique codes of the sMRI and fMRI modalities, respectively.

The variational loss aims to regularise the unique part of the latent space. The encoding distribution for the unique latent variables is $q(\mathbf{z}_u | \mathbf{x}_i)$, and the decoding distribution is $p(\mathbf{x}_i | \mathbf{z}_u)$. The VAE approximates the true posterior $p(\mathbf{z}_u | \mathbf{x}_i)$ with $q(\mathbf{z}_u | \mathbf{x}_i)$, employing the re-parameterisation trick for efficient optimisation.

The variational loss, represented as $\mathcal{L}_{var}^i$, is formulated as the Kullback-Leibler (KL) divergence $\mathcal{D}_{KL}(q(\mathbf{z}_u | \mathbf{x}_i) \| p(\mathbf{z}_u))$, which serves as a regularisation term to align $q(\mathbf{z}_u | \mathbf{x}_i)$ with the prior $p(\mathbf{z}_u)$, typically a standard Gaussian $\mathcal{N}(\mathbf{0}, \mathbf{I})$.

\begin{equation}
\mathcal{L}_{var}^i = \mathcal{D}_{KL}\left(q(\mathbf{z}_u | \mathbf{x}_i) \| p(\mathbf{z}_u)\right),
\end{equation}

The overall variational loss is the sum of the losses from both modalities:
\begin{equation}
\mathcal{L}_{var} = \mathcal{L}_{var}^1 + \mathcal{L}_{var}^2.
\end{equation}

This integration of variational loss facilitates learning rich and nuanced representations through a hybrid approach, combining the strengths of adversarial and probabilistic modelling for effective generative capabilities and structured latent space interpretation.

\subsubsection{Generic-Unique Distance Ratio Loss}
The distance ratio loss, $\mathcal{L}_{\text{Dist}}$, emphasises the disentanglement of latent variables by balancing the distances between the generic (shared) and unique components of the embeddings of two modalities. It is defined as follows:
\begin{equation}
\mathcal{L}_{\text{Dist}} = \frac{\mathcal{L}_{\text{Dist}}^{\text{Gen}}}{\mathcal{L}_{\text{Dist}}^{\text{Unq}}},
\end{equation}
where,
\begin{equation}
\mathcal{L}_{\text{Dist}}^{\text{Gen}} = \mathbb{E}_{\mathbf{x}_1, \mathbf{x}_2}\left\|\text{Gen}\left(\mathbf{Enc}_1\left(\mathbf{x}_1\right)\right) - \text{Gen}\left(\mathbf{Enc}_2\left(\mathbf{x}_2\right)\right)\right\|_2,
\end{equation}
and
\begin{equation}
\mathcal{L}_{\text{Dist}}^{\text{Unq}} = \mathbb{E}_{\mathbf{x}_1, \mathbf{x}_2}\left\|\text{Unq}\left(\mathbf{Enc}_1\left(\mathbf{x}_1\right)\right) - \text{Unq}\left(\mathbf{Enc}_2\left(\mathbf{x}_2\right)\right)\right\|_2.
\end{equation}

\subsubsection{Regression Loss}
The regression loss employs the L2 norm to measure the discrepancy between predicted and actual values:
\begin{equation}
\mathcal{L}_{\text{reg}} = \mathbb{E}_{\mathbf{x}_1, \mathbf{x}_2}\left\|y - \mathbf{P}\left(M\left(\mathbf{x}_1, \mathbf{x}_2\right)\right)\right\|_2.
\end{equation}

\subsubsection{Classification Loss}
The classification loss is formulated using Binary Cross-Entropy to evaluate the accuracy of binary classification tasks:
\begin{align}
\mathcal{L}_{\text{class}} = &y \log\left(\mathbf{C}\left(M\left(\mathbf{x}_1, \mathbf{x}_2\right)\right)\right) \nonumber \\
&+ (1-y) \log\left(1 -\mathbf{C}\left(M\left(\mathbf{x}_1, \mathbf{x}_2\right)\right)\right).
\end{align}

\subsubsection{Reconstruction Loss}
The reconstruction loss, designed for cross-modality reconstruction, excludes any computation from missing data to accommodate incomplete neuroimage datasets:
\begin{align}
\mathcal{L}_{\text{recon}} = &\sum_{i=1}^2 \sum_{j=1}^2 \mathbb{E}_{\mathbf{x}_i \sim P_d\left(\mathbf{x}_i\right)}\\ \bigg\|\mathbf{x}_i - \nonumber &\mathbf{Dec}_i\left(\text{Generic}\left(\mathbf{Enc}_j\left(\mathbf{x}_j\right)\right), \text{Unq}\left(\mathbf{Enc}_i\left(\mathbf{x}_i\right)\right)\right)\bigg\|_2.
\end{align}

\subsubsection{Full Objective}
The full objective function integrates all individual losses with corresponding trade-off parameters, aiming to optimize the model components cohesively:
\begin{equation}
\mathcal{L}_{\mathbf{D}} = \mathcal{L}_{\text{adv}},
\end{equation}

\begin{align}
\mathcal{L}_{\mathbf{Enc}_i, \mathbf{Dec}_i, \mathbf{P}, \mathbf{C}} = &\lambda_1 \mathcal{L}_{\text{reg}} + \lambda_2 \mathcal{L}_{\text{class}} + \lambda_3 \mathcal{L}_{\text{dist}} \nonumber \\
&+ \lambda_4 \mathcal{L}_{\text{recon}} + \lambda_5 \mathcal{L}_{\text{adv}} + \lambda_6 \mathcal{L}_{\text{var}},
\end{align}

where $\lambda_1, \lambda_2, \lambda_3, \lambda_4, \lambda_5$, and $\lambda_6$ are the trade-off parameters. The multitask adversarial variational autoencoder (M-AVAE) framework first updates the discriminator $\mathbf{D_z}$ to differentiate between true and generated samples, followed by updating the encoder $\mathbf{Enc}_i$, decoder $\mathbf{Dec}_i$, and predictor $\mathbf{P}$ based on the combined objective function, catering to both modality-specific and shared representations.

\begin{figure}[ht]
\centering
\includegraphics[width=0.5\textwidth]{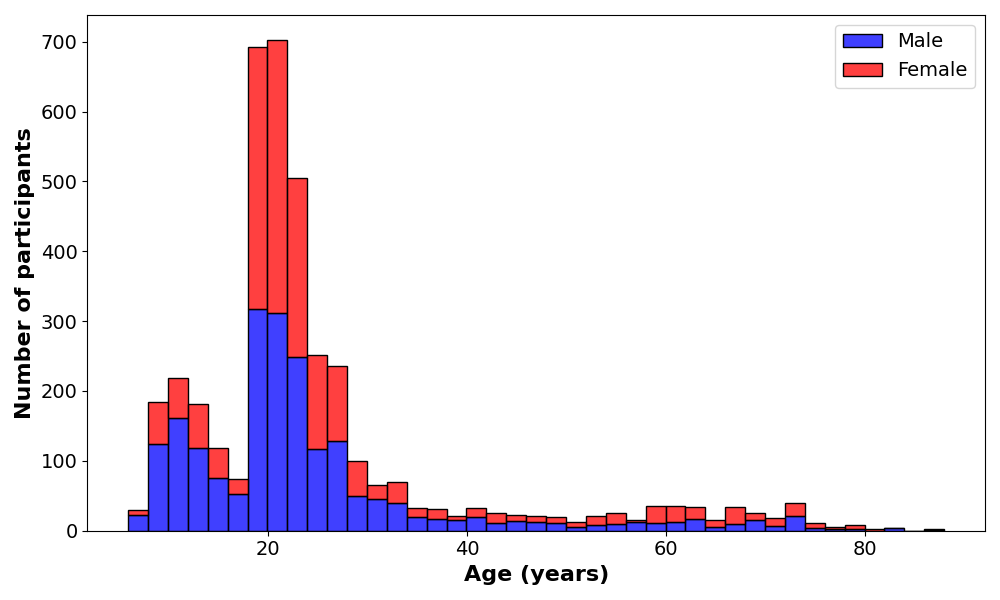}
\caption{Age and gender distribution across the training and validation sets of the OpenBHB dataset \cite{dufumier2022openbhb}.}
\label{data_distribution}
\end{figure}

\subsection{Datasets}

Our experiments leverage the OpenBHB dataset \cite{dufumier2022openbhb}, a comprehensive collection comprising 5,330 3D brain MRI scans from 71 different acquisition sites. Of these, 3,984 scans are publicly accessible, distributed across 3,227 training and 757 validation instances, as shown in Fig. \ref{data_distribution}. The latter includes 362 internal tests and 390 external tests. OpenBHB spans 10 datasets, featuring subjects of European-American, European, and Asian descent, ensuring a diverse range of genetic backgrounds. For our analysis, we focused on subsets containing both sMRI and fMRI scans, specifically using two datasets referenced in \cite{dataset1} and \cite{dataset2}, comprising 66 and 315 scans, respectively. These datasets were merged to facilitate our study, and we applied the preprocessing pipeline outlined in Section \ref{preprocessing} to both datasets.

\begin{figure}[ht]
\centering
\includegraphics[width=0.5\textwidth]{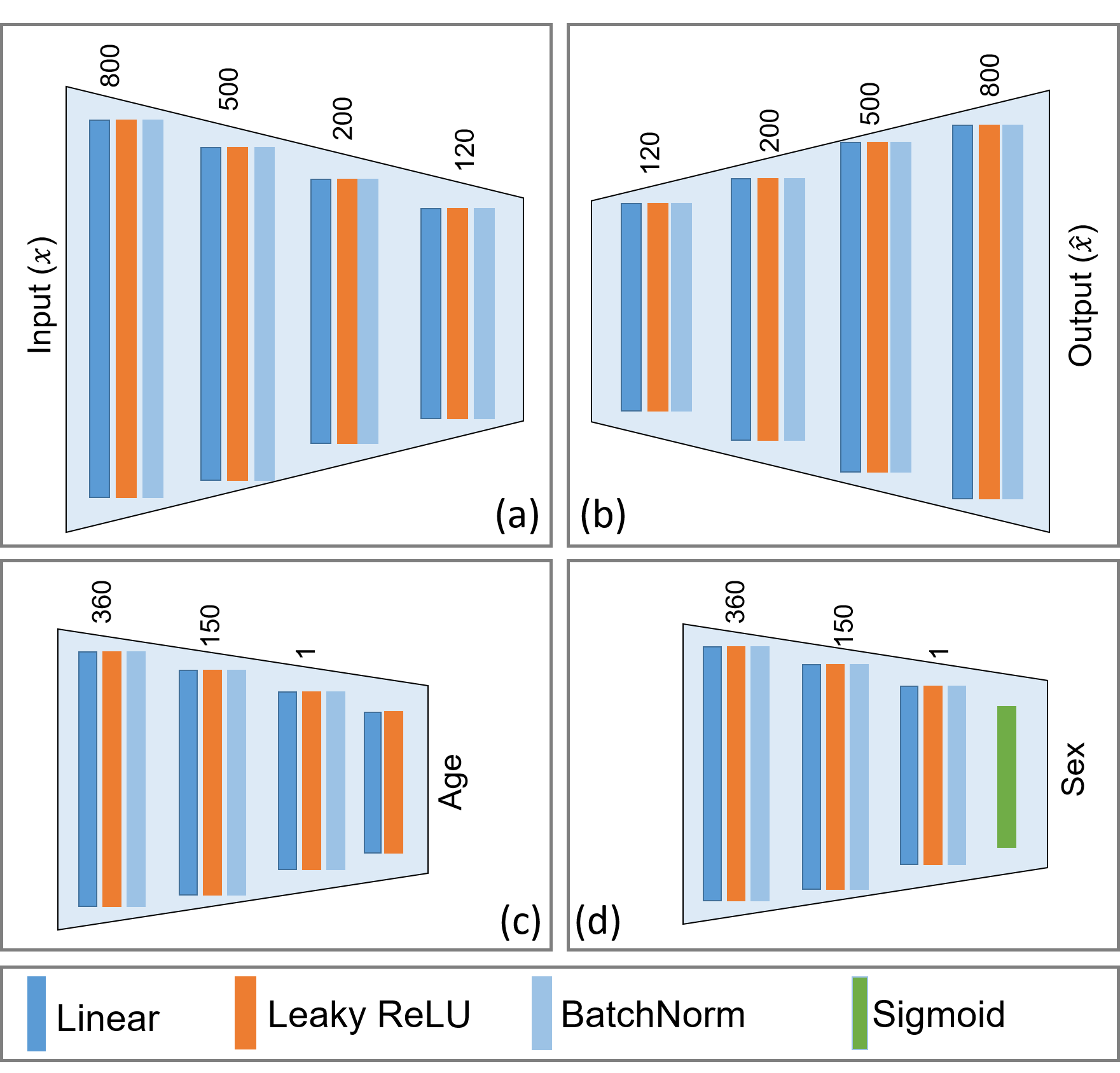}
\caption{Details of the various components of our proposed M-AVAE architecture. Subfigures (a), (b), (c), and (d) represent the architectural details of the encoder, decoder, age regressor, and sex classifier networks, respectively. Note that the discriminator shares the same architecture as the sex classifier.}
\label{backbone_net}
\end{figure}

\subsection{Model Architecture and Training Strategy}
The Multitask Adversarial Variational Autoencoder (M-AVAE) architecture, utilised in our experiments, is illustrated in Fig. \ref{backbone_net}. The batch size was established at 20, with the latent variable dimensionality set to 120. Furthermore, the dimensions of the generic and unique codes were determined to be 50 and 70, respectively, based on empirical tests. These dimensionality settings apply equally to all AAE models employed in our study, as represented in Fig. \ref{backbone_net}. For the purpose of single-task learning analysis, the gender classifier was deactivated. 
The training employed the Adam optimiser with an initial learning rate of 0.001. Learning rate adjustments were made by reducing it to one-quarter upon failing to observe performance enhancements on the validation set after nine epochs. To mitigate overfitting, an early stop mechanism was implemented. The models were developed using Keras with the TensorFlow back-end and trained on an NVIDIA RTX 4090 GPU. The number of parameters in our model is 1,144,372 and it took almost 12 and 1.5 hours to train and test the model, respectively.

\begin{table}[ht]
\caption{Summary of performance of various regression methods, including our proposed Multitask Adversarial Variational Autoencoder (M-AVAE), by presenting the mean and standard deviation for three metrics: Mean Absolute Error (MAE), Root Mean Square Error (RMSE), and Pearson Correlation Coefficient (PCC).}
\label{benchmarking_tab}
\centering
\begin{tabular}{
@{}>{\raggedright\arraybackslash}m{0.16\linewidth}
>{\centering\arraybackslash}m{0.2\linewidth}
>{\centering\arraybackslash}m{0.2\linewidth}
>{\centering\arraybackslash}m{0.2\linewidth}@{}}
\toprule
\textbf{\scriptsize Method Type} & \textbf{\scriptsize MAE} & \textbf{\scriptsize RMSE} & \textbf{\scriptsize PCC} \\
\midrule
\textbf{\scriptsize Model Agnostic} & & & \\
\scriptsize RF & \scriptsize 4.958 $\pm$ 3.000 & \scriptsize 5.795 $\pm$ 4.271 & \scriptsize 0.647 $\pm$ 0.290 \\
\scriptsize SVR & \scriptsize 4.458 $\pm$ 2.819 & \scriptsize 5.275 $\pm$ 4.039 & \scriptsize 0.685 $\pm$ 0.251 \\
\scriptsize GPR & \scriptsize 5.117 $\pm$ 2.830 & \scriptsize 5.848 $\pm$ 4.104 & \scriptsize 0.557 $\pm$ 0.183 \\
\scriptsize PLSR & \scriptsize 3.561 $\pm$ 1.993 & \scriptsize 4.081 $\pm$ 2.481 & \scriptsize 0.765 $\pm$ 0.172 \\
\addlinespace
\textbf{\scriptsize Model-based} & & & \\
\scriptsize MKL & \scriptsize 3.411 $\pm$ 2.147 & \scriptsize 4.031 $\pm$ 2.503 & \scriptsize 0.746 $\pm$ 0.160 \\
\scriptsize iMSF & \scriptsize 3.879 $\pm$ 2.533 & \scriptsize 4.633 $\pm$ 2.970 & \scriptsize 0.721 $\pm$ 0.165 \\
\addlinespace
\textbf{\scriptsize AAE-based} & & & \\
\scriptsize AAE & \scriptsize 3.126 $\pm$ 1.853 & \scriptsize 3.634 $\pm$ 2.177 & \scriptsize 0.796 $\pm$ 0.138 \\
\scriptsize M-AAE & \scriptsize 3.125 $\pm$ 1.976 & \scriptsize 3.697 $\pm$ 2.169 & \scriptsize 0.805 $\pm$ 0.121 \\
\scriptsize M-AVAE (Our proposed) & \scriptsize 2.773 $\pm$ 1.567 & \scriptsize 3.185 $\pm$ 1.901 & \scriptsize 0.824 $\pm$ 0.126 \\
\bottomrule
\end{tabular}
\end{table}

\begin{table}[ht]
\centering
\caption{Comparison of the brain age estimation performance of various methods, including our proposed M-AVAE model. The results presented for each method alongside the datasets used.}
\label{SOTA_comparison}
\begin{tabular}{
>{\hspace{0pt}}m{0.25\linewidth}
>{\hspace{0pt}}m{0.2\linewidth}
>{\centering\hspace{0pt}}m{0.1\linewidth}
>{\centering\hspace{0pt}}m{0.1\linewidth}
>{\centering\arraybackslash\hspace{0pt}}m{0.1\linewidth}
} 
\toprule
\textbf{\scriptsize Method} & \textbf{\scriptsize Dataset} & \textbf{\scriptsize MAE} & \textbf{\scriptsize RMSE} & \textbf{\scriptsize PCC} \\
\midrule
\scriptsize 3D-Peng \cite{peng2021accurate} & \scriptsize OASIS-3 \cite{lamontagne2019oasis} & \scriptsize 4.17 & \scriptsize 5.374 & \scriptsize ~- \\ 
\scriptsize Age-Net-Gender \cite{armanious2021age} & \scriptsize OASIS-3 \cite{lamontagne2019oasis}& \scriptsize 3.61 & \scriptsize 4.759 & \scriptsize - \\ 
\scriptsize CAE \cite{cai2023graph} & \scriptsize UKB \cite{sudlow2015uk} & \scriptsize 2.71 & \scriptsize 3.68 & \scriptsize 0.868 \\ 
\scriptsize Proposed M-AVAE & \scriptsize OpenBHB \cite{dufumier2022openbhb} & \scriptsize 2.773 & \scriptsize 3.185 & \scriptsize 0.824 \\
\bottomrule
\end{tabular}
\end{table}

\section*{Results and Discussion}
\label{res}
\subsection{Comparison With the State-of-the-Art Methods}
In this section, we evaluated the efficacy of our proposed Multitask Adversarial Variational Autoencoder (M-AVAE) model by comparing it against a diverse set of regression methodologies. These included four model-agnostic methods: Random Forest (RF), Support Vector Regression (SVR), Gaussian Process Regression (GPR), and Partial Least Squares Regression (PLSR); two model-based methods: Multiple Kernel Learning (MKL) \cite{wilson2019multiple} and Incomplete Multi-Source Fusion (iMSF) \cite{yuan2012multi}; and two approaches based on Adversarial Autoencoders (AAEs). The MKL method integrates dual kernels applied to sMRI and fMRI features to derive an optimal regression kernel, whereas iMSF focuses on learning shared feature sets with sparse regression across varying data source availability. The AAE methodologies take advantage of latent variables from sMRI and fMRI data for the prediction of brain age, with a variant that also predicts the gender of the patient (M-AAE).

To benchmark these models, we utilised three established metrics: Mean Absolute Error (MAE), Root Mean Square Error (RMSE), and Pearson's Correlation Coefficient (PCC). The comparative analysis, summarised in Table \ref{benchmarking_tab}, revealed that the traditional and model-based approaches yielded comparable results in age prediction. Specifically, the models evaluated showed MAEs ranging from 5.117 to 3.411 years, RMSE values approximately between 5.8 and 4 years, and PCCs of 0.55 to 0.74. In particular, AAE-based methods demonstrated superior performance, underscoring the advantages of adversarial learning and the use of age-related latent variables. The integration of sex information into the M-AAE model further improved its predictive accuracy. Ultimately, the M-AVAE model outperformed all the methodologies compared, showcasing the benefits of segregating latent variables derived from sMRI and fMRI into distinct components through adversarial and variational learning mechanisms.

Further analysis contrasted the M-AVAE model with previously published studies, as detailed in Table \ref{SOTA_comparison}. Among the selected benchmarks—3D-Peng \cite{peng2021accurate}, Age-Net-Gender \cite{armanious2021age}, and CAE \cite{cai2023graph}—our model surpassed in terms of MAE. Although the CAE model achieved marginally better MSE and PCC values, this can be attributed to its training on a significantly larger dataset. Our dataset comprised 381 sMRI and fMRI scans, while the OASIS-3 and UKB datasets contain 1,230 and 16,458 scans, respectively. This comparison highlights the efficacy of our hybrid regularization strategy, which combines adversarial and variational autoencoder techniques to generate more meaningful latent representations. This approach, coupled with the inclusion of sex information during training, significantly contributed to the superior performance of our proposed M-AVAE framework.

\begin{table}[ht]
  \centering
  \caption{Illustration of a detailed breakdown of Mean Absolute Error (MAE) results, measured in years and expressed as mean ± standard deviation, across various age groups: under 25 years (Y), 25 to 35 years, 35 to 45 years, and 45 to 55 years.}
  \label{results_breakdown_tab}
    \begin{tabular}{
    @{}>{\raggedright\arraybackslash}m{0.12\linewidth}
    >{\centering\arraybackslash}m{0.17\linewidth}
    >{\centering\arraybackslash}m{0.17\linewidth}
    >{\centering\arraybackslash}m{0.17\linewidth}
    >{\centering\arraybackslash}m{0.17\linewidth}@{}}
    \toprule
    \textbf{\scriptsize Method} & \textbf{\scriptsize $<$25Y} & \textbf{\scriptsize 25$\sim$35Y} & \textbf{\scriptsize 35$\sim$45Y} & \textbf{\scriptsize 45$\sim$55Y} \\
    \midrule
    \scriptsize RF & \scriptsize $4.78 \pm 2.93$ & \scriptsize $4.78 \pm 3.06$ & \scriptsize $6.20 \pm 2.68$ & \scriptsize $4.73 \pm 3.49$ \\
    \scriptsize SVR & \scriptsize $4.45 \pm 2.70$ & \scriptsize $5.09 \pm 3.17$ & \scriptsize $9.99 \pm 4.07$ & \scriptsize $8.16 \pm 2.76$ \\
    \scriptsize GPR & \scriptsize $4.73 \pm 2.67$ & \scriptsize $5.72 \pm 3.48$ & \scriptsize $7.47 \pm 2.05$ & \scriptsize $7.04 \pm 4.40$ \\
    \scriptsize PLSR & \scriptsize $3.37 \pm 1.94$ & \scriptsize $3.47 \pm 2.50$ & \scriptsize $5.72 \pm 4.45$ & \scriptsize $4.96 \pm 2.91$ \\
    \scriptsize MKL & \scriptsize $3.32 \pm 2.02$ & \scriptsize $3.60 \pm 2.53$ & \scriptsize $5.28 \pm 2.96$ & \scriptsize $4.97 \pm 3.70$ \\
    \scriptsize iMSF & \scriptsize $3.78 \pm 2.32$ & \scriptsize $4.05 \pm 2.50$ & \scriptsize $8.89 \pm 1.89$ & \scriptsize $6.35 \pm 4.40$ \\
    \scriptsize AAE & \scriptsize $3.00 \pm 1.65$ & \scriptsize $3.22 \pm 2.00$ & \scriptsize $3.87 \pm 1.61$ & \scriptsize $4.52 \pm 1.68$ \\
    \scriptsize M-AAE & \scriptsize $3.01 \pm 1.73$ & \scriptsize $3.03 \pm 2.22$ & \scriptsize $4.68 \pm 2.07$ & \scriptsize $3.90 \pm 1.91$ \\
    \scriptsize M-AVAE & \scriptsize $\mathbf{2.66 \pm 1.51}$ & \scriptsize $\mathbf{2.62 \pm 1.64}$ & \scriptsize $\mathbf{2.78 \pm 1.43}$ & \scriptsize $\mathbf{3.74 \pm 1.33}$ \\
    \bottomrule
    \end{tabular}
\end{table}

\begin{figure*}[ht]
\centering
\includegraphics[width=0.7\textwidth]{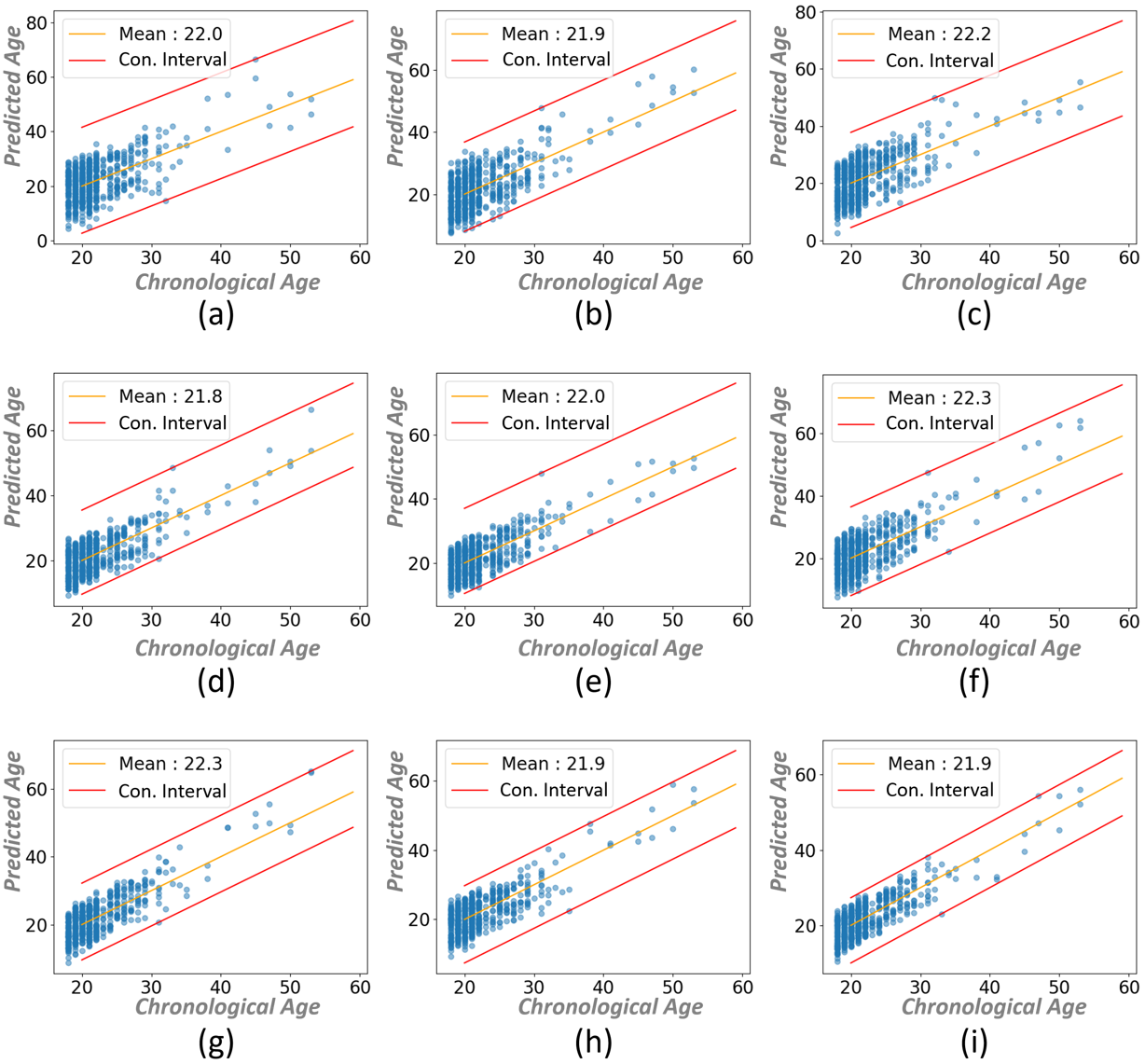}
\caption{Scatter plots illustrating the predicted age versus chronological age for nine different machine learning models. Subfigures (a), (b), (c), (d), (e), (f), (g), (h), and (i) represent the results from Random Forest (RF), Support Vector Regression (SVR), Gaussian Process Regression (GPR), Partial Least Squares Regression (PLSR), Multiple Kernel Learning (MKL) \cite{wilson2019multiple}, Incomplete Multi-Source Fusion (iMSF) \cite{yuan2012multi}, Adversarial Autoencoder (AAE), Multitask Adversarial Autoencoder (M-AAE), and the proposed Multitask Adversarial Variational Autoencoder (M-AVAE), respectively. The yellow line indicates the mean value, while the red lines represent the confidence intervals.} 
\label{results_plots}
\end{figure*}

\subsection{Robustness Analysis}

To assess the robustness of the proposed M-AVAE model, we conducted a 10-fold cross-validation to examine the consistency of the model's performance across the entire dataset. Scatter plots juxtaposing the predicted ages with the chronological ages, generated by M-AVAE and other models evaluated under identical experimental conditions, are shown in Figure \ref{results_plots}. This comparison illuminates the relative performance and robustness of each model.

The analysis reveals that the model-agnostic methods (RF, SVR, GPR, and PLSR) and the model-based techniques (MKL and iMSF) exhibit comparable mean performance metrics. However, PLSR and iMSF demonstrate enhanced confidence intervals, indicating more reliable performance predictions. Among these, AAE-based models significantly outperform in terms of confidence intervals, highlighting the robustness of AAE-based approaches. In particular, the multitask AAE model surpasses the single task AAE variant, underscoring the value of incorporating gender information through multitask learning. Our M-AVAE model achieves superior performance, exemplifying the effectiveness of our approach in creating a disentangled latent space. This is achieved by enforcing three distinct prior distributions, leveraging both adversarial and variational losses, which signifies the robustness and adaptability of the M-AVAE model.

Further examination of prediction accuracy across various age groups is detailed in Table \ref{results_breakdown_tab}. The analysis presents differential performance across age segments. In particular, all models exhibit improved accuracy in younger age groups (below 25 years), with MAE values ranging from 4.78 to 2.66 years. However, performance disparities become more pronounced in older age segments, particularly within the $35\sim45$ and $45\sim55$ year ranges, where performance metrics vary significantly—highlighting a decline in model accuracy for older individuals. Despite these variations, AAE-based models maintain consistent performance, reinforcing their robustness. Within this category, the Multitask AAE model consistently outperforms its single-task counterpart, further validating the benefit of integrating gender information through multitask learning. Ultimately, the M-AVAE model distinguishes itself by delivering the most consistent and accurate predictions across all age groups, affirming the efficacy of our method in generating a disentangled latent space through the application of adversarial and variational principles.

\subsection{Multi-Modality Analysis}

\begin{figure*}[!t]
\centering
\includegraphics[width=0.7\textwidth]{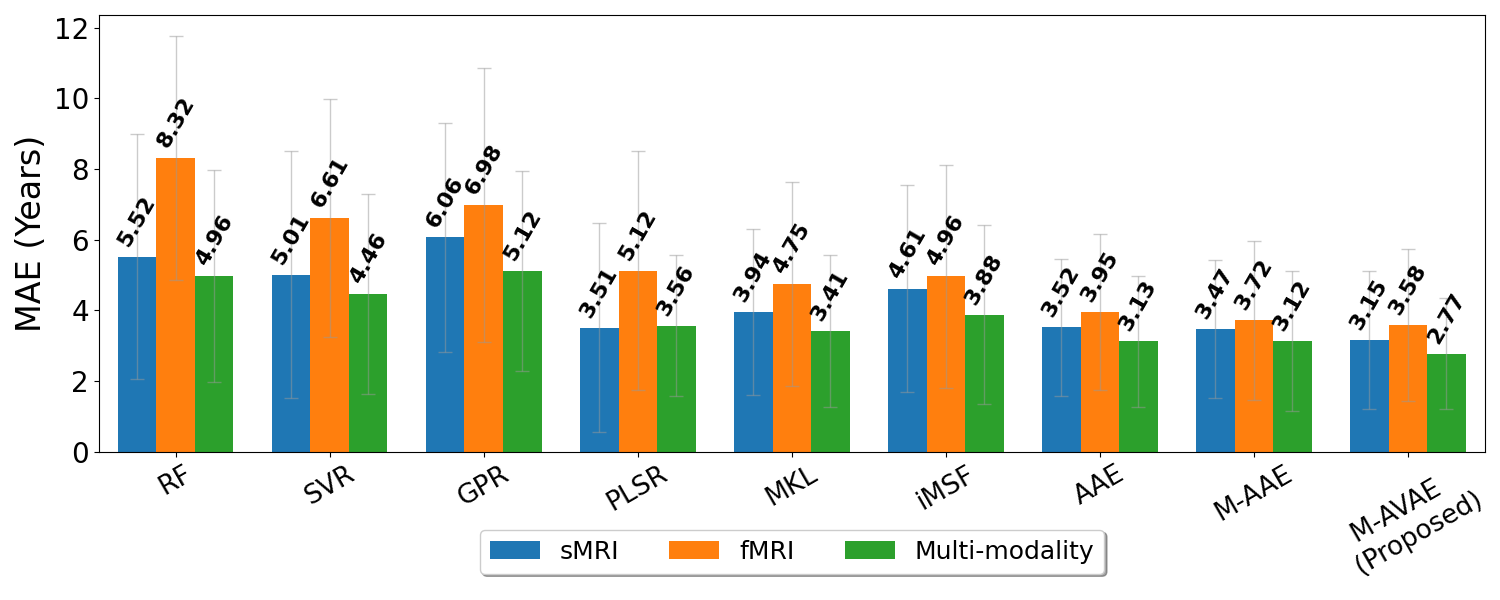}
\caption{The bar chart compares the mean absolute error (MAE) in years of various multimodal and unimodal regression methods for brain age prediction. }
\label{results_bar_chart}
\end{figure*}

The impact of multi-modality fusion on model performance was scrutinized by comparing unimodal (sMRI or fMRI alone) and multimodal (combined sMRI and fMRI) data approaches, as depicted in Figure \ref{results_bar_chart}. The analysis revealed that unimodal models trained exclusively on fMRI yielded the least favourable results, with those trained on sMRI performing moderately better. In contrast, models that used a multimodal strategy, incorporating sMRI and fMRI scans, demonstrated superior performance, underscoring the benefits of such an integrative approach.

Specifically, the Mean Absolute Error (MAE) for age prediction using only sMRI data ranged from 6.06 to 3.15 years. In contrast, models that relied on fMRI data alone showed MAEs between 8.32 and 3.58 years. Using multimodal data, MAEs improved, covering 4.82 to 2.77 years. This incremental improvement from sMRI-based unimodal to multimodal strategies was consistently observed in all models, including the proposed M-AVAE. This trend suggests that while traditional multimodal fusion methods might introduce additional noise from fMRI data, thus reducing accuracy compared to using only sMRI data, the M-AVAE approach effectively mitigates this issue. In particular, M-AVAE reduced MAE from 3.15 years with sMRI data alone to 2.77 years by employing the multimodal strategy, highlighting the model's ability to integrate multimodal information.

\section*{Conclusions}
\label{con}
In this study, we introduced the Multi-Task Adversarial Variational Autoencoder (M-AVAE), a framework designed for brain age estimation, with potential applications in the evolving landscape of the Metaverse and healthcare. Our approach utilises multimodal Magnetic Resonance Imaging (MRI) data, combining structural and functional MRI to create a more comprehensive latent representation. Through the integration of techniques such as cross-reconstruction, adversarial and variational learning, and the introduction of a distance ratio loss, M-AVAE aims to disentangle latent variables into shared and modality-specific components. This process facilitates more effective data fusion and improves prediction accuracy, especially with the incorporation of sex information as an auxiliary task, acknowledging biological factors in age estimation. The performance of M-AVAE was evaluated on the OpenBHB dataset, where it demonstrated promising results, achieving a mean absolute error of 2.77 years in predicting brain age. Although these findings suggest that the M-AVAE framework can provide significant improvements over some traditional methods, further validation is necessary across different datasets and scenarios to fully understand its broader applicability. We believe that the insights gained from this work could contribute to the development of more refined diagnostic tools in virtual healthcare environments, although further research and exploration are essential to realise their full potential.


\begin{thebibliography}{10}
\providecommand{\url}[1]{#1}
\csname url@samestyle\endcsname
\providecommand{\newblock}{\relax}
\providecommand{\bibinfo}[2]{#2}
\providecommand{\BIBentrySTDinterwordspacing}{\spaceskip=0pt\relax}
\providecommand{\BIBentryALTinterwordstretchfactor}{4}
\providecommand{\BIBentryALTinterwordspacing}{\spaceskip=\fontdimen2\font plus
\BIBentryALTinterwordstretchfactor\fontdimen3\font minus \fontdimen4\font\relax}
\providecommand{\BIBforeignlanguage}[2]{{%
\expandafter\ifx\csname l@#1\endcsname\relax
\typeout{** WARNING: IEEEtran.bst: No hyphenation pattern has been}%
\typeout{** loaded for the language `#1'. Using the pattern for}%
\typeout{** the default language instead.}%
\else
\language=\csname l@#1\endcsname
\fi
#2}}
\providecommand{\BIBdecl}{\relax}
\BIBdecl

\bibitem{wang2022development}
G.~Wang, A.~Badal, X.~Jia, J.~S. Maltz, K.~Mueller, K.~J. Myers, C.~Niu, M.~Vannier, P.~Yan, Z.~Yu \emph{et~al.}, ``Development of metaverse for intelligent healthcare,'' \emph{Nature Machine Intelligence}, vol.~4, no.~11, pp. 922--929, 2022.

\bibitem{laubenbacher2024digital}
R.~Laubenbacher, B.~Mehrad, I.~Shmulevich, and N.~Trayanova, ``Digital twins in medicine,'' \emph{Nature Computational Science}, vol.~4, no.~3, pp. 184--191, 2024.

\bibitem{katsoulakis2024digital}
E.~Katsoulakis, Q.~Wang, H.~Wu, L.~Shahriyari, R.~Fletcher, J.~Liu, L.~Achenie, H.~Liu, P.~Jackson, Y.~Xiao \emph{et~al.}, ``Digital twins for health: a scoping review,'' \emph{NPJ Digital Medicine}, vol.~7, no.~1, p.~77, 2024.

\bibitem{el2019dtwins}
A.~El~Saddik, H.~Badawi, R.~A.~M. Velazquez, F.~Laamarti, R.~G. Diaz, N.~Bagaria, and J.~S. Arteaga-Falconi, ``Dtwins: A digital twins ecosystem for health and well-being,'' \emph{IEEE COMSOC MMTC Commun. Front}, vol.~14, no.~2, pp. 39--43, 2019.

\bibitem{bib_1}
J.~H. Cole and K.~Franke, ``Predicting age using neuroimaging: Innovative brain ageing biomarkers,'' \emph{Trends Neurosci.}, vol.~40, no.~12, pp. 681--690, Dec 2017.

\bibitem{bib_2}
J.~H. Cole, R.~Leech, and D.~J. Sharp, ``Prediction of brain age suggests accelerated atrophy after traumatic brain injury,'' \emph{Ann. Neurol.}, vol.~77, no.~4, pp. 571--581, Apr 2015.

\bibitem{li2021_alzheimer}
B.~Li \emph{et~al.}, ``Identifying individuals with alzheimer’s disease-like brains based on structural imaging in the human connectome project aging cohort,'' \emph{Human Brain Mapping}, vol.~42, no.~17, pp. 5535--5546, 2021.

\bibitem{bib_3}
I.~Nenadi{\'c}, M.~Dietzek, K.~Langbein, H.~Sauer, and C.~Gaser, ``Brainage score indicates accelerated brain aging in schizophrenia, but not bipolar disorder,'' \emph{Psychiatry Res., Neurimag.}, vol. 266, pp. 86--89, Aug 2017.

\bibitem{tanveer2023deep}
M.~Tanveer, M.~Ganaie, I.~Beheshti, T.~Goel, N.~Ahmad, K.-T. Lai, K.~Huang, Y.-D. Zhang, J.~Del~Ser, and C.-T. Lin, ``Deep learning for brain age estimation: A systematic review,'' \emph{Information Fusion}, vol.~96, pp. 130--143, 2023.

\bibitem{bib_9}
F.~Liem and et~al., ``Predicting brain-age from multimodal imaging data captures cognitive impairment,'' \emph{NeuroImage}, vol. 148, pp. 179--188, Mar 2017.

\bibitem{franke2013gender}
K.~Franke, M.~Ristow, and C.~Gaser, ``Gender-specific effects of health and lifestyle markers on individual brainage,'' in \emph{2013 International Workshop on Pattern Recognition in Neuroimaging}.\hskip 1em plus 0.5em minus 0.4em\relax IEEE, 2013, pp. 94--97.

\bibitem{sanford2022sex}
N.~Sanford, R.~Ge, M.~Antoniades, A.~Modabbernia, S.~S. Haas, H.~C. Whalley, L.~Galea, S.~G. Popescu, J.~H. Cole, and S.~Frangou, ``Sex differences in predictors and regional patterns of brain age gap estimates,'' \emph{Human Brain Mapping}, vol.~43, no.~15, pp. 4689--4698, 2022.

\bibitem{he2022global}
S.~He, P.~E. Grant, and Y.~Ou, ``Global-local transformer for brain age estimation,'' \emph{IEEE Trans. Med. Imaging}, vol.~41, no.~1, pp. 213--224, Jan 2022.

\bibitem{he2022deep}
S.~He, Y.~Feng, P.~E. Grant, and Y.~Ou, ``Deep relation learning for regression and its application to brain age estimation,'' \emph{IEEE Trans. Med. Imaging}, vol.~41, no.~9, pp. 2304--2317, Sep 2022.

\bibitem{cheng2021brain}
J.~Cheng and et~al., ``Brain age estimation from mri using cascade networks with ranking loss,'' \emph{IEEE Trans. Med. Imaging}, vol.~40, no.~12, pp. 3400--3412, Dec 2021.

\bibitem{armanious2021age}
K.~Armanious and et~al., ``Age-net: An mri-based iterative framework for brain biological age estimation,'' \emph{IEEE Trans. Med. Imaging}, vol.~40, no.~7, pp. 1778--1791, Jul 2021.

\bibitem{zhang2022robust}
Z.~Zhang and et~al., ``Robust brain age estimation based on smri via nonlinear age-adaptive ensemble learning,'' \emph{IEEE Trans. Neural Syst. Rehabil. Eng.}, vol.~30, pp. 2146--2156, 2022.

\bibitem{liu2023risk}
W.~Liu, Q.~Dong, S.~Sun, J.~Shen, K.~Qian, and B.~Hu, ``Risk prediction of alzheimer’s disease conversion in mild cognitive impaired population based on brain age estimation,'' \emph{IEEE Trans. Neural Syst. Rehabil. Eng.}, vol.~31, pp. 2468--2476, 2023.

\bibitem{dular2024base}
L.~Dular and Z.~Spiclin, ``Base: Brain age standardized evaluation,'' \emph{NeuroImage}, vol. 285, p. 120469, Jan 2024.

\bibitem{wang2023skewed}
H.~Wang, M.~S. Treder, D.~Marshall, D.~K. Jones, and Y.~Li, ``A skewed loss function for correcting predictive bias in brain age prediction,'' \emph{IEEE Trans. Med. Imaging}, vol.~42, no.~6, pp. 1577--1589, Jun 2023.

\bibitem{mouches2022multimodal}
P.~Mouches, M.~Wilms, D.~Rajashekar, S.~Langner, and N.~D. Forkert, ``Multimodal biological brain age prediction using magnetic resonance imaging and angiography with the identification of predictive regions,'' \emph{Human Brain Mapping}, vol.~43, no.~8, pp. 2554--2566, 2022.

\bibitem{cai2023graph}
H.~Cai, Y.~Gao, and M.~Liu, ``Graph transformer geometric learning of brain networks using multimodal mr images for brain age estimation,'' \emph{IEEE Trans. Med. Imaging}, vol.~42, no.~2, pp. 456--466, Feb 2023.

\bibitem{hu2020disentangled}
D.~Hu and et~al., ``Disentangled-multimodal adversarial autoencoder: Application to infant age prediction with incomplete multimodal neuroimages,'' \emph{IEEE Trans. Med. Imaging}, vol.~39, no.~12, pp. 4137--4149, Dec 2020.

\bibitem{bedel2023bolt}
H.~A. Bedel, I.~Sivgin, O.~Dalmaz, S.~U. Dar, and T.~{\c{C}}ukur, ``Bolt: Fused window transformers for fmri time series analysis,'' \emph{Medical image analysis}, vol.~88, p. 102841, 2023.

\bibitem{bedel2023dreamr}
H.~A. Bedel and T.~{\c{C}}ukur, ``Dreamr: Diffusion-driven counterfactual explanation for functional mri,'' \emph{arXiv preprint arXiv:2307.09547}, 2023.

\bibitem{jirsaraie2023systematic}
R.~J. Jirsaraie, A.~J. Gorelik, M.~M. Gatavins, D.~A. Engemann, R.~Bogdan, D.~M. Barch, and A.~Sotiras, ``A systematic review of multimodal brain age studies: Uncovering a divergence between model accuracy and utility,'' \emph{Patterns}, vol.~4, no.~4, 2023.

\bibitem{bib_20}
T.~Baltrusaitis, C.~Ahuja, and L.-P. Morency, ``Multimodal machine learning: A survey and taxonomy,'' \emph{IEEE Trans. Pattern Anal. Mach. Intell.}, vol.~41, no.~2, pp. 423--443, Feb 2019.

\bibitem{liu2018learn}
K.~Liu, Y.~Li, N.~Xu, and P.~Natarajan, ``Learn to combine modalities in multimodal deep learning,'' \emph{arXiv preprint arXiv:1805.11730}, 2018.

\bibitem{ullah2020hybrid}
Z.~Ullah, M.~U. Farooq, S.-H. Lee, and D.~An, ``A hybrid image enhancement based brain mri images classification technique,'' \emph{Medical hypotheses}, vol. 143, p. 109922, 2020.

\bibitem{farooq2023residual}
M.~U. Farooq, Z.~Ullah, and J.~Gwak, ``Residual attention based uncertainty-guided mean teacher model for semi-supervised breast masses segmentation in 2d ultrasonography,'' \emph{Computerized Medical Imaging and Graphics}, vol. 104, p. 102173, 2023.

\bibitem{kanwal2023mask}
M.~Kanwal, M.~M. Ur~Rehman, M.~U. Farooq, and D.-K. Chae, ``Mask-transformer-based networks for teeth segmentation in panoramic radiographs,'' \emph{Bioengineering}, vol.~10, no.~7, p. 843, 2023.

\bibitem{usman2024intelligent}
M.~Usman, A.~Rehman, S.~Masood, T.~M. Khan, and J.~Qadir, ``Intelligent healthcare system for iomt-integrated sonography: Leveraging multi-scale self-guided attention networks and dynamic self-distillation,'' \emph{Internet of Things}, vol.~25, p. 101065, 2024.

\bibitem{usman2023mesaha}
M.~Usman, A.~Rehman, A.~Shahid, S.~Latif, S.~S. Byon, S.~H. Kim, T.~M. Khan, and Y.~G. Shin, ``Mesaha-net: Multi-encoders based self-adaptive hard attention network with maximum intensity projections for lung nodule segmentation in ct scan,'' \emph{arXiv preprint arXiv:2304.01576}, 2023.

\bibitem{usman2024meds}
M.~Usman, A.~Rehman, A.~Shahid, S.~Latif, and Y.-G. Shin, ``Meds-net: Multi-encoder based self-distilled network with bidirectional maximum intensity projections fusion for lung nodule detection,'' \emph{Engineering Applications of Artificial Intelligence}, vol. 129, p. 107597, 2024.

\bibitem{rehman2023selective}
A.~Rehman, M.~Usman, A.~Shahid, S.~Latif, and J.~Qadir, ``Selective deeply supervised multi-scale attention network for brain tumor segmentation,'' \emph{Sensors}, vol.~23, no.~4, p. 2346, 2023.

\bibitem{iqbal2023ldmres}
S.~Iqbal, T.~M. Khan, S.~S. Naqvi, A.~Naveed, M.~Usman, H.~A. Khan, and I.~Razzak, ``Ldmres-net: a lightweight neural network for efficient medical image segmentation on iot and edge devices,'' \emph{IEEE Journal of Biomedical and Health Informatics}, 2023.

\bibitem{usman2023deha}
M.~Usman and Y.-G. Shin, ``Deha-net: a dual-encoder-based hard attention network with an adaptive roi mechanism for lung nodule segmentation,'' \emph{Sensors}, vol.~23, no.~4, p. 1989, 2023.

\bibitem{usman2022dual}
M.~Usman, A.~Rehman, A.~M. Saleem, R.~Jawaid, S.-S. Byon, S.-H. Kim, B.-D. Lee, M.-S. Heo, and Y.-G. Shin, ``Dual-stage deeply supervised attention-based convolutional neural networks for mandibular canal segmentation in cbct scans,'' \emph{Sensors}, vol.~22, no.~24, p. 9877, 2022.

\bibitem{ullah2023ssmd}
Z.~Ullah, M.~Usman, S.~Latif, A.~Khan, and J.~Gwak, ``Ssmd-unet: semi-supervised multi-task decoders network for diabetic retinopathy segmentation,'' \emph{Scientific Reports}, vol.~13, no.~1, p. 9087, 2023.

\bibitem{latif2018automating}
S.~Latif, M.~Asim, M.~Usman, J.~Qadir, and R.~Rana, ``Automating motion correction in multishot mri using generative adversarial networks,'' \emph{arXiv preprint arXiv:1811.09750}, 2018.

\bibitem{lee2021evaluation}
M.~S. Lee, Y.~S. Kim, M.~Kim, M.~Usman, S.~S. Byon, S.~H. Kim, B.~I. Lee, and B.-D. Lee, ``Evaluation of the feasibility of explainable computer-aided detection of cardiomegaly on chest radiographs using deep learning,'' \emph{Scientific reports}, vol.~11, no.~1, p. 16885, 2021.

\bibitem{latif2018mobile}
S.~Latif, M.~Y. Khan, A.~Qayyum, J.~Qadir, M.~Usman, S.~M. Ali, Q.~H. Abbasi, and M.~A. Imran, ``Mobile technologies for managing non-communicable diseases in developing countries,'' in \emph{Mobile applications and solutions for social inclusion}.\hskip 1em plus 0.5em minus 0.4em\relax IGI Global, 2018, pp. 261--287.

\bibitem{ullah2023mtss}
Z.~Ullah, M.~Usman, and J.~Gwak, ``Mtss-aae: Multi-task semi-supervised adversarial autoencoding for covid-19 detection based on chest x-ray images,'' \emph{Expert Systems with Applications}, vol. 216, p. 119475, 2023.

\bibitem{ullah2023densely}
Z.~Ullah, M.~Usman, S.~Latif, and J.~Gwak, ``Densely attention mechanism based network for covid-19 detection in chest x-rays,'' \emph{Scientific Reports}, vol.~13, no.~1, p. 261, 2023.

\bibitem{usman2017using}
M.~Usman, S.~Latif, and J.~Qadir, ``Using deep autoencoders for facial expression recognition,'' in \emph{2017 13th International Conference on Emerging Technologies (ICET)}.\hskip 1em plus 0.5em minus 0.4em\relax IEEE, 2017, pp. 1--6.

\bibitem{ullah2022cascade}
Z.~Ullah, M.~Usman, M.~Jeon, and J.~Gwak, ``Cascade multiscale residual attention cnns with adaptive roi for automatic brain tumor segmentation,'' \emph{Information sciences}, vol. 608, pp. 1541--1556, 2022.

\bibitem{usman2020retrospective}
M.~Usman, S.~Latif, M.~Asim, B.-D. Lee, and J.~Qadir, ``Retrospective motion correction in multishot mri using generative adversarial network,'' \emph{Scientific reports}, vol.~10, no.~1, p. 4786, 2020.

\bibitem{usman2020volumetric}
M.~Usman, B.-D. Lee, S.-S. Byon, S.-H. Kim, B.-i. Lee, and Y.-G. Shin, ``Volumetric lung nodule segmentation using adaptive roi with multi-view residual learning,'' \emph{Scientific Reports}, vol.~10, no.~1, p. 12839, 2020.

\bibitem{latif2018cross}
S.~Latif, A.~Qayyum, M.~Usman, and J.~Qadir, ``Cross lingual speech emotion recognition: Urdu vs. western languages,'' in \emph{2018 International conference on frontiers of information technology (FIT)}.\hskip 1em plus 0.5em minus 0.4em\relax IEEE, 2018, pp. 88--93.

\bibitem{latif2018phonocardiographic}
S.~Latif, M.~Usman, R.~Rana, and J.~Qadir, ``Phonocardiographic sensing using deep learning for abnormal heartbeat detection,'' \emph{IEEE Sensors Journal}, vol.~18, no.~22, pp. 9393--9400, 2018.

\bibitem{latif2020leveraging}
S.~Latif, M.~Usman, S.~Manzoor, W.~Iqbal, J.~Qadir, G.~Tyson, I.~Castro, A.~Razi, M.~N.~K. Boulos, A.~Weller \emph{et~al.}, ``Leveraging data science to combat covid-19: A comprehensive review,'' \emph{IEEE Transactions on Artificial Intelligence}, vol.~1, no.~1, pp. 85--103, 2020.

\bibitem{farooq2023dc}
M.~U. Farooq, Z.~Ullah, A.~Khan, and J.~Gwak, ``Dc-aae: dual channel adversarial autoencoder with multitask learning for kl-grade classification in knee radiographs,'' \emph{Computers in Biology and Medicine}, vol. 167, p. 107570, 2023.

\bibitem{urbanowicz2018relief}
R.~J. Urbanowicz, M.~Meeker, W.~La~Cava, R.~S. Olson, and J.~H. Moore, ``Relief-based feature selection: Introduction and review,'' \emph{Journal of biomedical informatics}, vol.~85, pp. 189--203, 2018.

\bibitem{kawakubo2012rapid}
H.~Kawakubo and H.~Yoshida, ``Rapid feature selection based on random forests for high-dimensional data,'' \emph{Expert Syst. Appl}, vol.~40, pp. 6241--6252, 2012.

\bibitem{makhzani2015adversarial}
A.~Makhzani, J.~Shlens, N.~Jaitly, I.~Goodfellow, and B.~Frey, ``Adversarial autoencoders,'' \emph{arXiv preprint arXiv:1511.05644}, 2015.

\bibitem{kingma2013auto}
D.~P. Kingma and M.~Welling, ``Auto-encoding variational bayes,'' \emph{arXiv preprint arXiv:1312.6114}, 2013.

\bibitem{dufumier2022openbhb}
B.~Dufumier, A.~Grigis, J.~Victor, C.~Ambroise, V.~Frouin, and E.~Duchesnay, ``Openbhb: a large-scale multi-site brain mri data-set for age prediction and debiasing,'' \emph{NeuroImage}, vol. 263, p. 119637, 2022.

\bibitem{dataset1}
A.~Sunavsky and J.~Poppenk, ``Neuroimaging predictors of creativity in healthy adults,'' \emph{Neuroimage}, vol. 206, p. 116292, 2020.

\bibitem{dataset2}
S.~A. Nastase, Y.-F. Liu, H.~Hillman, A.~Zadbood, L.~Hasenfratz, N.~Keshavarzian, J.~Chen, C.~J. Honey, Y.~Yeshurun, M.~Regev \emph{et~al.}, ``The “narratives” fmri dataset for evaluating models of naturalistic language comprehension,'' \emph{Scientific data}, vol.~8, no.~1, p. 250, 2021.

\bibitem{peng2021accurate}
H.~Peng, W.~Gong, C.~F. Beckmann, A.~Vedaldi, and S.~M. Smith, ``Accurate brain age prediction with lightweight deep neural networks,'' \emph{Medical image analysis}, vol.~68, p. 101871, 2021.

\bibitem{lamontagne2019oasis}
P.~J. LaMontagne, T.~L. Benzinger, J.~C. Morris, S.~Keefe, R.~Hornbeck, C.~Xiong, E.~Grant, J.~Hassenstab, K.~Moulder, A.~G. Vlassenko \emph{et~al.}, ``Oasis-3: longitudinal neuroimaging, clinical, and cognitive dataset for normal aging and alzheimer disease,'' \emph{MedRxiv}, pp. 2019--12, 2019.

\bibitem{sudlow2015uk}
C.~Sudlow, J.~Gallacher, N.~Allen, V.~Beral, P.~Burton, J.~Danesh, P.~Downey, P.~Elliott, J.~Green, M.~Landray \emph{et~al.}, ``Uk biobank: an open access resource for identifying the causes of a wide range of complex diseases of middle and old age,'' \emph{PLoS medicine}, vol.~12, no.~3, p. e1001779, 2015.

\bibitem{wilson2019multiple}
C.~M. Wilson, K.~Li, X.~Yu, P.-F. Kuan, and X.~Wang, ``Multiple-kernel learning for genomic data mining and prediction,'' \emph{BMC bioinformatics}, vol.~20, pp. 1--7, 2019.

\bibitem{yuan2012multi}
L.~Yuan, Y.~Wang, P.~M. Thompson, V.~A. Narayan, and J.~Ye, ``Multi-source learning for joint analysis of incomplete multi-modality neuroimaging data,'' in \emph{Proceedings of the 18th ACM SIGKDD international conference on Knowledge discovery and data mining}, 2012, pp. 1149--1157.

\bibitem{islam2023revealing}
M.~T. Islam, Z.~Zhou, H.~Ren, M.~B. Khuzani, D.~Kapp, J.~Zou, L.~Tian, J.~C. Liao, and L.~Xing, ``Revealing hidden patterns in deep neural network feature space continuum via manifold learning,'' \emph{Nature Communications}, vol.~14, no.~1, p. 8506, 2023.

\end{thebibliography}

\end{document}